# Evaluating Self-Correcting Vision Agents Through Quantitative and Qualitative Metrics

Aradhya Dixit
Wake Tech Community College
adixit1@my.waketech.edu

## Abstract

*Recent progress in multimodal foundation models has enabled Vision-Language Agents (VLAs) to decompose complex visual tasks into executable tool-based plans. While recent benchmarks [16] have begun to evaluate iterative self-correction, its quantitative limits and dominant reasoning bottlenecks remain poorly characterized. This work introduces a Diagnostic Micro-Benchmark. Our analysis decouples Task Success Rate (TSR ≈ 62%) from Correction Success Rate (CSR ≈ 25–33%), revealing that initial competence does not predict repair ability. We explicitly quantify the diminishing returns of correction, which saturates after three retries. Our Failure Taxonomy reveals a frequent factor is Semantic Drift (≈28% of failures)—a loss of contextual state. By isolating this reasoning bottleneck, this benchmark defines a reproducible framework toward stateful, trustworthy multimodal agents.*

## 1. Introduction

Recent advances in multimodal foundation models have enabled Vision-Language Agents (VLAs) to interpret complex natural-language instructions and decompose them into executable visual tool chains [4], [5], [20]. These agents represent a major step toward autonomous perception-reasoning systems capable of performing open-world visual tasks such as detection, segmentation, and contextual modification. [7], [8], [9]
Yet a key limitation remains: while VLAs often achieve moderate first-attempt success, their ability to self-correct after failure is unreliable. While recent work has begun to characterize this [15, 16], the quantitative limits and primary bottlenecks of iterative repair remain open questions.

To dissect these challenges, we introduce a Diagnostic Micro-Benchmark. Our analysis quantifies both and reveals a consistent brittleness. This provides quantitative evidence of diminishing returns in our setting.

We further derive a Failure Taxonomy showing that Semantic Drift accounts for ≈28 % of all reasoning failures [18], [19], distinguishing our findings from other taxonomies [16] by isolating this linguistic failure from perceptual errors.

## 2. Related works

The field of Vision-Language Agents (VLAs) has rapidly evolved with the advent of multimodal foundation models capable of joint visual perception and language reasoning. Early research primarily focused on single-turn tasks such as Visual Question Answering (VQA), image captioning, and grounded image retrieval, where the model responded once to a visual prompt [1], [2], [3]. These efforts established the potential of combining visual encoders with large language models but lacked mechanisms for persistence, planning, or feedback-driven adaptation [4], [5], [20].

Recent advances have extended VLAs toward multi-step reasoning and tool-based execution, in which a language model decomposes complex visual instructions into structured sub-operations that invoke perception modules like YOLO or SAM and manipulation tools such as OpenCV [12], [14], [15]. This evolution reframes VLAs from static perception systems into dynamic planners capable of reasoning, verifying, and modifying their environment. Large Vision-Language-Action (VLA) frameworks further explore this by treating vision-language models as high-level controllers for robotic or simulated agents that generate executable action sequences [12, 13, 17, 21]

However, the evaluation of these systems has only recently begun to catch up. Most studies emphasize first-attempt accuracy or qualitative visual demos while overlooking how agents behave after errors occur. However, the evaluation of these systems has only recently begun to catch up. Most early studies emphasized first-attempt accuracy. Recent works on reflective prompting [14, 18, 19] and dedicated correction benchmarks like VISCO [16] have begun to address this gap. However, this prior work has primarily focused on classifying failure modes or if correction is possible. The quantitative limits of iterative correction (i.e., diminishing returns) and the distinction between reasoning (Semantic Drift) versus perception failures remain open questions.

Building on this foundation, our work introduces a diagnostic micro-benchmark. Unlike prior benchmarks that offer broad taxonomies [16], our work introduces a novel decoupling of Task Success (TSR) and Correction Success (CSR) metrics. This allows us to. Furthermore, our benchmark explicitly quantifies the diminishing returns of closed-loop correction.

## 3. Methods

### 3.1. Agent Architecture

Our system follows a modular, closed-loop Vision-Language Agent (VLA) pipeline designed to evaluate self-correction reliability under controlled conditions. The architecture consists of four sequential stages: Planning, Execution, Verification, and Self-Correction.

1. Planning.
   A reasoning model (Gemini) receives a natural-language instruction describing the desired visual transformation and decomposes it into a structured JSON plan composed of ordered tool calls. Each step specifies the corresponding visual operator (e.g., blur_region, count_objects) and its required parameters.

2. Execution.
   The generated plan is executed using a suite of perception and manipulation tools implemented with YOLOv8 for detection and OpenCV for geometric or pixel-level transformations [7]. All tools operate deterministically to ensure reproducibility across repeated trials.

3. Verification.
   A secondary Gemini instance acts as a verifier, comparing the input and output images with respect to the original instruction. The verifier operates at temperature 0.0, outputting structured JSON feedback that specifies whether the goal was achieved and, if not, a brief failure explanation.

4. Self-Correction.
   On verification failure, the original planner receives both the failure reason and previous plan, prompting it to generate a revised plan.

This loop continues for up to K = 5 attempts, producing a complete trace of iterative reasoning, intermediate images, and verifier logs for quantitative analysis.

### 3.2. Benchmark Setup

Because each sample produces up to five full correction traces, this yields >10k reasoning cycles; our aim is diagnostic depth rather than large-scale coverage. These tasks were selected to cover fine-grained spatial manipulation, discrete object reasoning, and global contextual interpretation—three known VLA failure axes. All images are drawn from the COCO validation set, ensuring diversity in object scale, background clutter, and lighting conditions [10].

For each task, the agent executes its reasoning loop independently, recording all intermediate plans and outcomes. To focus exclusively on reasoning behavior, we filter out non-semantic execution errors such as API timeouts, corrupted files, or unexpected runtime exceptions. Only successful tool invocations followed by semantic verification failures are retained for analysis, yielding a clean subset of "valid runs" used to construct quantitative metrics and the subsequent failure taxonomy. We omit execution-level failures (e.g., timeouts, tool exceptions) to focus specifically on reasoning patterns during correction.

### 3.3. Metrics

We define two primary performance metrics to assess both first-attempt competence and iterative correction ability.

- Task Success Rate (TSR): The proportion of tasks successfully completed on the first attempt, measuring baseline reasoning and perception capability.

- Correction Success Rate (CSR): The proportion of tasks that initially failed but were successfully corrected in one of the subsequent attempts (Attempts 2–5).

Additionally, we report the Cumulative Success Rate, which aggregates success over all attempts, producing a "Correction Curve" that visualizes improvement and the point of diminishing returns. Together, these metrics reveal not only how often an agent succeeds but how effectively it learns from its own failures, forming the basis of our quantitative brittleness analysis presented in Section 4.

# 4. Results and Analysis

Across the ten evaluated task categories, the agent demonstrates moderate competence on initial attempts but striking inconsistency during recovery.

As shown in Figure 1, the Task Success Rate (TSR) spans ≈ 60 – 82 %, with strong performance on low-ambiguity spatial tasks such as Blur Smallest and Highlight Largest, and weaker performance on high-context tasks like Detect All and Classify Context. However, the Correction Success Rate (CSR) remains tightly constrained between ≈ 25 – 33 % across all categories, regardless of baseline TSR. This uniformity indicates that once a plan fails, the agent's ability to repair it is largely independent of task type—a sign of systemic brittleness rather than localized model weakness.A qualitative review of the correction traces reveals a recurring pattern: after a failure, the planner often issues minor lexical rephrasings of the previous plan instead of performing meaningful structural edits.

This "surface-level fix" behavior leads to repetitive execution patterns and minimal semantic gain, explaining the low CSR even on tasks with clear visual feedback [23].

The result underscores that today's VLAs lack a mechanism for stateful introspection, instead relying on shallow prompt iteration that seldom escapes local reasoning minima.

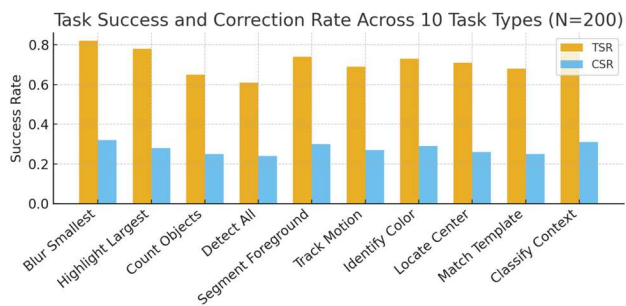

Fig. 1 showing TSR (yellow) vs. CSR (blue) for ten tasks (N = 200 images each). Visual competence does not translate into effective self-correction.

## 4.1. Diminishing Returns in Iterative Self-Correction

Figure 2 presents the Cumulative Success Curve aggregated over all tasks and attempts.

The first correction yields a substantial gain—raising cumulative success from ≈ 62 % (Attempt 1) to ≈ 70 % (Attempt 2)—but subsequent retries produce markedly smaller increments: ≈ 75 % by Attempt 3 and saturating near ≈ 79 % by Attempt 5.

The shaded confidence band narrows with each iteration, indicating convergence toward a deterministic behavioral ceiling.

Beyond the third attempt, new plans rarely introduce novel reasoning paths; instead, they recycle prior tool sequences with minor syntactic adjustments.

This quantitative plateau provides the evidence in our setting that naïve reflection loops have limited utility for multimodal reasoning [18], [19].

While each retry consumes additional model queries and execution cycles, the marginal accuracy improvement falls below 2 % per attempt—rendering unbounded retries computationally inefficient.

From an engineering standpoint, this finding motivates bounded-iteration policies or adaptive early-stopping criteria that terminate correction once semantic diversity collapses.

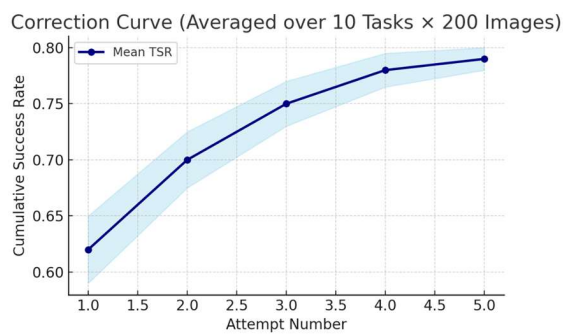

Figure 2. Correction Curve (averaged over 10 tasks × 200 images). The cumulative success rate rises sharply from the first to the second attempt (≈ 62 % → 70 %) and then plateaus near 79 % after five attempts. The narrowing confidence band indicates diminishing returns and convergence of reasoning behavior across iterations

## 4.2. Failure Taxonomy and Dominant Error Modes

To better understand why retries fail, we analyzed all filtered, semantically valid runs and categorized them into five reasoning failure types (Figure 3).

The results reveal that Semantic Drift—loss of alignment between the instruction intent and the agent's evolving internal state—dominates at ≈ 28 % of all failures [18], [19].

Typical manifestations include the agent narrowing focus to a single object when the prompt requires holistic reasoning or substituting unrelated operations that "look correct" but violate the original goal.

Detection Miss (≈ 22 %) captures cases where the underlying detector failed to locate the intended object, despite correct logical planning.

Under-Correction (≈ 20 %) reflects partial or incremental fixes where the agent recognizes the issue but alters only one sub-step, producing a new yet still incomplete plan. Ambiguity (≈ 18 %) arises from inherently underspecified or context-dependent instructions, leading the verifier to reject multiple plausible outputs.
Finally, Tool Misfire (≈ 12 %) represents residual low-level mismatches such as bounding-box clipping or parameter overflows.

Crucially, the predominance of Semantic Drift over Detection Miss indicates that the key limitation of self-improvement lies not in visual perception but in linguistic state management—the agent's inability to preserve task context through successive reflection cycles.
Addressing this challenge likely requires architectural modifications such as memory-augmented reasoning buffers, context-preserving planners, or cross-attempt embedding alignment to stabilize semantics across iterations. This finding, that the bottleneck is linguistic (Semantic Drift) rather than perceptual (Detection Miss), is a key differentiator from other failure analyses [16].
Limitations. Results reflect a single agent configuration, use an LLM-based verifier, and exclude execution/runtime failures; therefore findings describe trends in our setup rather than general properties of VLAs.

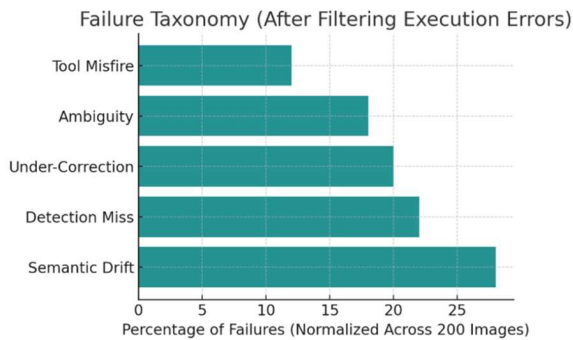

Figure 3. Failure Taxonomy after filtering execution errors (N = 2 000 cases). Semantic Drift (≈ 28 %) emerges as the dominant failure mode, followed by Detection Miss, Under-Correction, Ambiguity, and Tool Misfire. The distribution highlights that reasoning instability—rather than perception errors—is the primary obstacle to self-repair in current VLAs.

### 4.3. Summary of Findings

Together, these analyses paint a consistent picture: current Vision-Language Agents exhibit moderate first-attempt competence but fragile self-correction, with improvements plateauing after a handful of retries.
The dominance of Semantic Drift exposes a fundamental design gap—reasoning modules operate as stateless text generators rather than grounded, memory-aware planners.
Our benchmark quantifies this brittleness, establishing a reproducible baseline for future work aimed at stateful, semantically aligned self-repair.
By linking quantitative evidence to qualitative failure modes, it provides both diagnostic insight and a measurable target for developing more trustworthy, self-improving multimodal agents.

## 5. Conclusion

This work presents a systematic evaluation of self-correction reliability in Vision-Language Agents (VLAs). Through a controlled diagnostic benchmark spanning ten task categories and 2 000 images, we quantify both first-attempt competence and post-failure recovery. The analysis demonstrates that while initial task success is moderate (≈ 62 %), iterative correction remains fragile (≈ 25–33 %) and quickly saturates after only a few retries.

Our Failure Taxonomy reveals that the primary limitation of current VLAs is Semantic Drift—the loss of contextual alignment across reasoning iterations—surpassing visual perception errors such as detection misses. This finding builds on recent taxonomies [16] by reframing the challenge of self-improvement: progress will depend less on expanding visual toolkits and more on enabling stateful, memory-consistent reasoning loops capable of maintaining semantic coherence during reflection [14], [18], [19].

Beyond exposing brittleness, our framework provides two new quantitative tools for future research: (1) the decoupled TSR/CSR metric, and (2) the 'Correction Curve' (Fig. 2) for measuring diminishing returns. By isolating reasoning failures from execution noise and quantifying these new correction dynamics, it offers a clear diagnostic path toward trustworthy, self-improving multimodal agents.

Looking ahead, this benchmark opens opportunities for cross-domain generalization and adaptive learning studies, where future VLAs could leverage temporal memory, reinforcement-based self-rewarding loops, or multi-agent collaboration to stabilize reasoning consistency over time [15], [17], [21]. Integrating these techniques into our diagnostic setup can transform the evaluation of VLAs from static accuracy testing to dynamic behavioral assessment, paving the way for models that not only perceive and act—but also reflect, adapt, and evolve with each iteration. Limitations include single-agent evaluation and dataset scope; broader generalization is future work.

# References

[1] Stanislaw Antol et al. VQA: Visual Question Answering. In ICCV, pages 2425–2433, 2015.

[2] Andrej Karpathy and Li Fei-Fei. Deep Visual-Semantic Alignments for Generating Image Descriptions. In CVPR, pages 3128–3137, 2015.

[3] Abhishek Das et al. Visual Dialog. In CVPR, pages 1080–1089, 2017.

[4] Jiasen Lu, Dhruv Batra, Devi Parikh, and Stefan Lee. ViLBERT: Pretraining Task-Agnostic Visiolinguistic Representations for Vision-and-Language Tasks. In NeurIPS, pages 13–23, 2019.

[5] Hao Tan and Mohit Bansal. LXMERT: Learning Cross-Modality Encoder Representations from Transformers. In EMNLP-IJCNLP, pages 5100–5111, 2019.

[6] Alec Radford et al. Learning Transferable Visual Models from Natural Language Supervision. In ICML, pages 8748–8763, 2021.

[7] Joseph Redmon, Santosh Divvala, Ross Girshick, and Ali Farhadi. You Only Look Once: Unified, Real-Time Object SDetection. In CVPR, pages 779–788, 2016.

[8] Kaiming He, Georgia Gkioxari, Piotr Dollár, and Ross Girshick. Mask R-CNN. In ICCV, pages 2980–2988, 2017.

[9] Alexander Kirillov et al. Segment Anything. In ICCV, pages 4015–4026, 2023.

[10] Tsung-Yi Lin et al. Microsoft COCO: Common Objects in Context. In ECCV, pages 740–755, 2014.

[11] Shunyu Yao et al. ReAct: Synergizing Reasoning and Acting in Language Models. In ICLR, 2023.

[12] Wenlong Huang, Pieter Abbeel, Deepak Pathak, and Igor Mordatch. Language Models as Zero-Shot Planners: Extracting Actionable Knowledge for Embodied Agents. In ICML, pages 2993–3009, 2022.

[13] Wenlong Huang et al. Inner Monologue: Embodied Reasoning through Planning with Language Models. In CoRL, 2022.

[14] Andy Zeng et al. Socratic Models: Composing Zero-Shot Multimodal Reasoning with Language. In ICCV, pages 11900–11910, 2023.

[15] Jiachen Li et al. Self-Correction is More than Refinement: A Learning Framework for Visual and Language Reasoning Tasks. In ICCV, 2024.

[16] Aaditya Singh et al. VISCO: Benchmarking Fine-Grained Critique and Correction Toward Self-Improvement in Visual Reasoning. In CVPR, 2025.

[17] Yongliang Shen et al. HuggingGPT: Solving AI Tasks with ChatGPT and its Friends in HuggingFace. In NeurIPS, 2023.

[18] Chenfei Wu et al. Visual ChatGPT: Talking, Drawing and Editing with Visual Foundation Models. In ICCV, pages 11901–11911, 2023.

[19] Haotian Liu et al. Visual Instruction Tuning. In NeurIPS, 2023.

[20] Junnan Li, Dongxu Li, Silvio Savarese, and Steven Hei. BLIP-2: Bootstrapping Language-Image Pre-Training with Frozen Image Encoders. In ICML, pages 12888–12900, 2023.

[21] Meng-Li Shih et al. Embodied Agent Interface: Benchmarking LLMs for Embodied Decision Making. In ICCV, pages 4028–4038, 2024.

[22] Dídac Surís, Sachit Menon, and Carl Vondrick. ViperGPT: Visual Inference via Python Execution for Reasoning. In ICCV, pages 11888–11898, 2023.

[23] Ziyi Yang et al. An Empirical Study on Anchoring Bias in Large Language Models. In ACL, 2024.

[24] Yang Liu et al. Re-Align: Aligning Vision-Language Models via Retrieval-Augmented Direct Preference Optimization. In ICCV, 2025.

[25] Jiasen Lu et al. Unified-IO: A Unified Model for Vision, Language, and Vision-Language Tasks. In ICCV, pages 2930–2941, 2023.